\documentclass[cameraready]{Interspeech}

\title{Automatic estimation of verbal fluency index in people with Motor Neuron Disease using ASR alignment and pause modelling
}

\author[affiliation={1}]{Bahman}{Mirheidari}
\author[affiliation={2}]{Leslie}{Ing}
\author[affiliation={2}]{Daniel}{Blackburn}
\author[affiliation={3,4}]{Sharon} {Abrahams}
\author[affiliation={2}]{Christopher}{McDermott}
\author[affiliation={1}]{Heidi}{Christensen}

\address{
  $^{1}$School of Computer Science, University of Sheffield, UK;
  $^{2}$Sheffield Institute for Translational Neuroscience (SITraN); $^{3}$Department of Psychology, School of Philosophy, Psychology and Language Sciences, University of Edinburgh, UK; $^{4}$Euan MacDonald Centre for Motor Neuron Disease Research, University of Edinburgh, UK}

\email{b.mirheidari@sheffield.ac.uk,heidi.christensen@sheffield.ac.uk}


\usepackage{comment}
\usepackage{bm}
\usepackage{tabularx}
\usepackage{booktabs}

\usepackage[utf8]{inputenc}

\newif\ifanonymous
\anonymoustrue

\begin{document}

\maketitle

\begin{abstract}   

Monitoring cognitive impairment (CI) in motor neuron disease (MND) is essential for timely treatment and care, yet challenging due to co-occurring speech difficulties. The Edinburgh Cognitive and Behavioural ALS Screen (ECAS) provides a robust metric for CI assessment, with the Verbal Fluency Index (VFI) a central element. Building on recent advances in automated speech analysis, this study proposes a system for estimating VFI. It leverages a unique MND dataset and combines ASR (WhisperX) and VAD (Silero) with refined timestamping to predict the VFI and extract several clinically interpretable measures. Our approach outperformed systems based on traditional acoustic features and self-supervised embeddings, evaluated using multiple regression algorithms. Clinically inspired features consistently outperformed the other sets, with the best models achieving strong results (P-words: R² 0.9, NRMSE 0.05; S-words: R² 0.8, NRMSE 0.08), demonstrating the feasibility of automated VFI estimation.
\end{abstract}

\section{Introduction}

Motor Neuron Disease (MND) is an umbrella term covering a set of progressive neurodegenerative diseases characterised by motor neuron degeneration affecting both limb and bulbar function, and Amyotrophic Lateral Sclerosis (ALS) is the most common form of MND \cite{marangi2024als}. Over 5,000 people are living with MND in the UK \cite{mndassociation2025facts}, with an incidence of approximately 5.7 per 100,000 people in England annually \cite{burchardt2022analysis}. Up to 50\% of ALS patients develop cognitive impairment (CI), characterised by executive dysfunction and verbal fluency impairment, with 15\% developing frontotemporal dementia (FTD) \cite{rutkove2023digital,abrahams2023neuropsychological}. The Edinburgh Cognitive and Behavioural ALS Screen (ECAS, total score: 136) \cite{Abrahams2015ECAS} is a multi-domain neuropsychological screening tool designed to detect cognitive and behavioural impairment of people with ALS, demonstrating high sensitivity and specificity for frontotemporal dysfunction \cite{niven2015validation}. Assessing cognitive function in this population is inherently challenging, as the presence of speech and motor deficits may confound traditional performance metrics. The ECAS addresses this issue in its verbal fluency subtest through the Verbal Fluency Index (VFI). Patients complete a one-minute fluency test: they must say as many words as possible beginning with a target letter, excluding proper nouns like countries or names. 
It is an articulation-speed-adjusted index that provides a purer index of cognitive retrieval speed than raw word counts, which can be conflated with impairment when slowed speech is mistaken for executive dysfunction \cite{abrahams2000verbal}.

VFI calculation requires two phases: a \textit{generation phase}, in which participants produce words within one minute, followed by a \textit{controlled reading phase}, where clinicians identify correct words (excluding repetitions, hesitations, and proper nouns) and then measure the time taken for the participant to read them aloud. Finally, the resulting VFI value is an estimate of the average time taken to think of each word. In the ECAS, the score is normalised against pre-defined thresholds to determine the corresponding verbal fluency score, which ranges from 0 to 12. This is a very time-consuming process, and Automatic VFI scoring from audio recordings could offer significant benefits: it reduces the need for clinician supervision, enables scalable online remote assessment, and reduces burden on both patients and the NHS. Since people with MND require routine cognitive assessment, which is not currently widely available in the UK, one of our project objectives is to facilitate automated assessment to address this gap.

In this study, we develop a pipeline using state-of-the-art automatic speech recognition (ASR) and voice activity detection (VAD) to estimate VFI while also generating precise timing, pause, and word-level information. The pipeline produces clinician-interpretable outputs, including lists of correct words and errors, that can be shared directly with health care professionals. This approach removes the stress of in-person consultations and reduces bias and subjectivity in identifying correct words and timing. Crucially, we demonstrate that the generation phase alone may suffice for producing reliable VFI scores, potentially eliminating the need for the controlled reading phase and its associated timing. We demonstrate the automated system's ability to estimate VFI and further analyse the results on MND patients with dysarthria and those with typical speech. This study uses a novel dataset collected from individuals with MND. Expert clinicians calculated ground-truth ECAS scores, including detailed domain-specific subscores and the VFI metrics.

Section \ref{sec:background} reviews recent work on automatic speech assessment and clinical scoring for people with MND. Section \ref{sec:pipeline} then describes our proposed pipeline. Section \ref{sec:setup} outlines the experimental setup, followed by Section \ref{sec:results}, which presents our results and findings. Finally, Section \ref{sec:conclusions} concludes the paper.

\section{Related work}
\label{sec:background}
Several recent studies have investigated speech-based assessment for individuals with dysarthria and ALS, utilising both traditional techniques and advanced AI models. However, due to persistent data limitations in the medical domain, conventional approaches remain relevant and practically useful. For instance, traditional acoustic features remain valuable for assessing ALS-related dysarthria. Speaking rate and pause frequency, for example, are sensitive to disease progression, often more so than standard clinical scales  \cite{neumann2024multimodal}. Similarly, spectral features are associated with both motor and CI \cite{desilva2024variability}. Acoustic features outperform text for multi-class ALS dysarthria severity classification, though both perform similarly in binary tasks \cite{pakhomov2015using}. Recent studies, however, demonstrate that combining both modalities achieves optimal performance \cite{ys2025comparison}. 


The automation of verbal fluency assessment has been advanced considerably through Automatic Speech Recognition (ASR) technologies. Early work by \cite{konig2018fully} established the feasibility of this approach, extracting features like switching rates from semantic fluency tasks in cognitively impaired populations and achieving near-perfect agreement with manual scoring  ($r \approx 0.9$). Subsequent innovations have addressed specific limitations of this initial approach \cite{troger2023post} and enhanced temporal accuracy through post-processing of ASR output. In parallel, methodological advances have expanded the scope of automated assessment for people with dysarthria. \cite{xiong2024improving} employed multi-task learning to overcome data limitations in dysarthria detection (differentiating dysarthric from typical speech), and \cite{wang2024automatic} showed that transformer models like ALS Longitudinal Speech Transformer (ALST) can capture temporal speech dynamics to estimate disease progression with 91.0\% AUC. Most comprehensively, \cite{neumann2024multimodal} integrated multimodal biomarkers, combining acoustic, facial, and linguistic features, to detect bulbar decline earlier than conventional clinical scales.

Advances in speech-based assessment, from traditional acoustic analysis to sophisticated AI techniques including multi-task learning, transformer architectures, and multimodal biomarkers, now enable earlier detection of decline than conventional clinical scales. The existing studies have pursued related but distinct objectives. Some have focused on assessing dysarthric speech \cite{YeoCKC23, Fougeron2022}, while others have automated verbal fluency scoring by counting valid words in general or clinical populations. To date, no study has specifically targeted the automatic estimation of the ECAS VFI in people with MND. This represents a critical gap, as the VFI was explicitly designed to address the confounding of motor and CI in this population, making it well-suited to automated approaches that capture both speech timing and content. Targeting clinical analysis of verbal fluency, our pipeline therefore prioritises precise automatic timing information and granular word-level assessment over advanced speech and text-based foundation models. In our proposed approach, we introduce an automatic pipeline that operates solely on the verbal fluency generation phase, eliminating the need for the controlled reading phase. Implementing the controlled reading phase presents a significant technical challenge: it imposes a significant burden on both patients and clinicians while being susceptible to inter-rater variability, and requires real-time, highly accurate ASR capable of perfectly identifying words and displaying them on screen.

\section{Automatic VFI scoring pipeline}  
\label{sec:pipeline}
\begin{figure}[t]
\captionsetup{font=footnotesize}

\footnotesize  
  \hspace{-0.7cm}
  \includegraphics[width=1.15\linewidth]{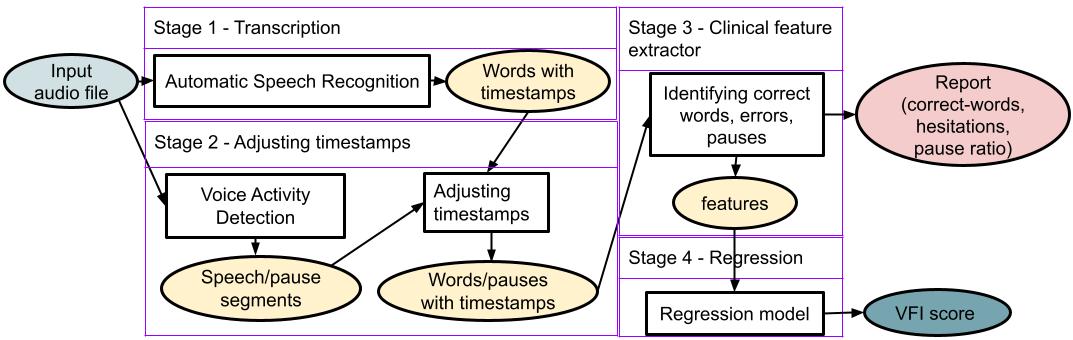}
  \caption{Automatic VFI scoring pipeline.}
  \label{fig:pipeline} 
   \vspace{-0.8cm}
\end{figure}

ASR timestamps alone are insufficient for VFI estimation, as accurate speech/pause segmentation boundaries are essential. The mismatch is clear: ASR models trained on thousands of hours of continuous speech struggle with the isolated word production typical of verbal fluency. Word boundaries produced by ASR systems lack the necessary precision, and crucially, pauses occurring between and after words are frequently misidentified or omitted. So we need an automatic approach to adjust boundaries. To address this limitation, we apply voice activity detection (VAD) and combine its outputs with the ASR timestamps to accurately generate the word and pause boundaries. In our proposed approach, we introduce an automatic pipeline that operates solely on the verbal fluency generation phase, eliminating the need for the controlled reading phase. The pipeline, illustrated in Figure \ref{fig:pipeline}, automatically scores the Verbal Fluency Index (VFI) through four stages: 1) Transcription (converts audio to text with word-level timestamps), 2) Adjusting timestamps (refines accuracy for words and pause boundaries), 3) Clinical feature extractor (on our target tasks, S-words and P-words, verifies word's correctness by excluding proper nouns, repetitions, hesitations, and errors, for correct-words plus gaps between them extracts statistical features), 4) Regression (trains and applies models to estimate VFI). 

\begin{figure}[t]
\captionsetup{font=footnotesize}
\footnotesize
  \centering 
  \includegraphics[width=0.925\linewidth]{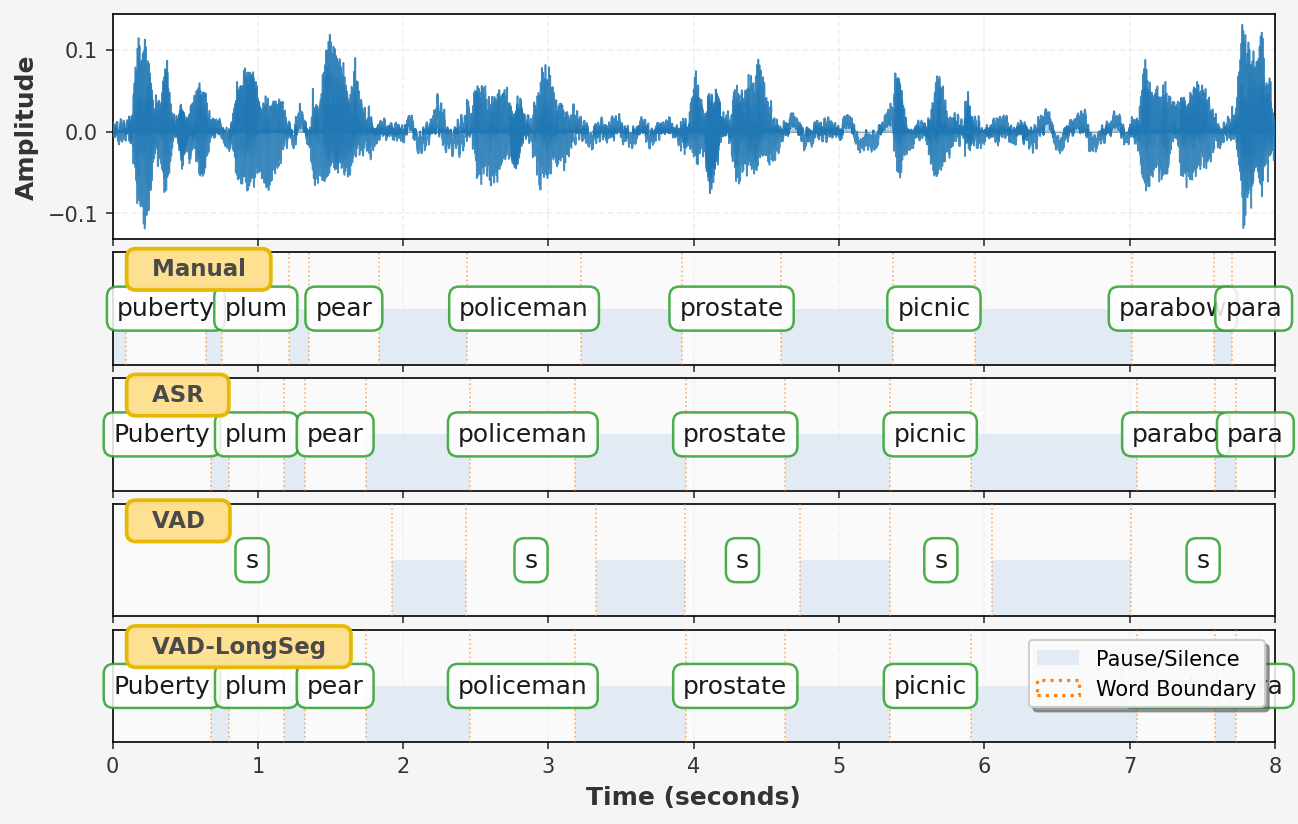}
    \includegraphics[width=0.925\linewidth]{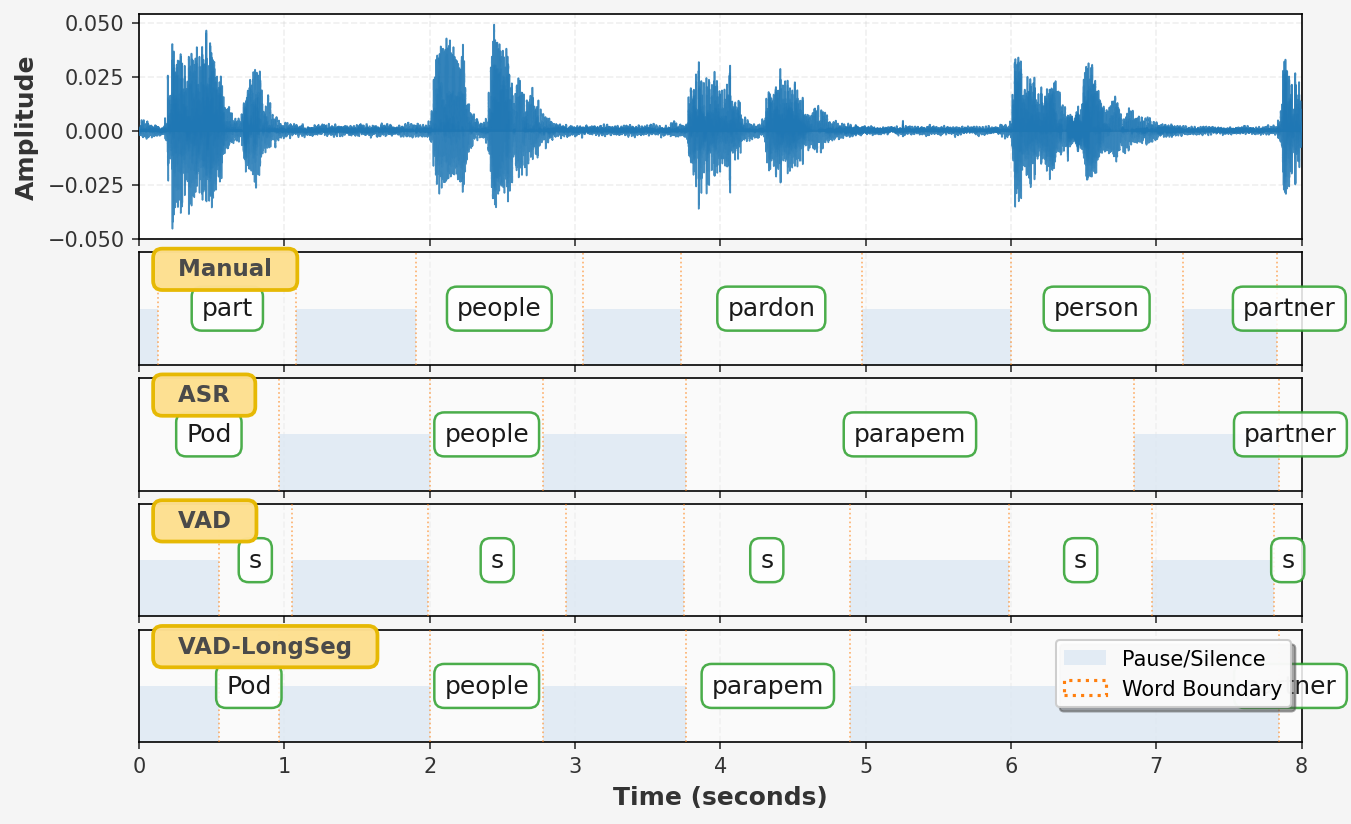}
  \caption{Manual alignment vs ASR, VAD and adjustments (top: typical speech, bottom: dysarthric).} 
  \label{fig:alignment} 
  \vspace{-0.3cm}
\end{figure}

Figure~\ref{fig:alignment} compares manual segmentation for typical speech (top) versus dysarthric speech (bottom) against: (i) raw ASR output, (ii) raw VAD output, and (iii) adjusting strategies like Last-Segment, First-Segment, and Longest-Segment (to resolve cases where multiple VAD speech segments aligned with a single ASR segment, selecting either the first, last, or longest segment). In the figure, for example, "puberty," "plum," and "pear" do not match the initial VAD segment. A matching procedure identifies overlapping regions, and multiple VAD segments may correspond to a single ASR word, as with "parapem," which overlaps two VAD segments. Outcomes include a single contiguous segment or multiple overlapping regions, requiring a selection strategy like choosing the longest remaining segment (the first candidate). ASR performance is often compromised with dysarthric speech, causing errors like "pod" for "part" and "parapem" for "pardon person." Background noise also necessitates a robust VAD approach for reliable speech detection. Because our features rely on precise timestamps for correctly produced words and the gaps between them, VAD outputs, which provide only speech-pause boundaries, are inadequate, requiring the use of adjusted boundaries.


\vspace{-0.3cm}
\section{Experimental setup}
\label{sec:setup}

\subsection{Dataset}
The dataset has been collected as part of an ongoing project, {CognoMND}, a version of the {CognoSpeak}\footnote{https://www.cognospeak.co.uk} platform adapted specifically for individuals with MND. {CognoSpeak} is a web-based tool that enables remote data collection from people with various medical conditions.  Participants completed a verbal fluency test, naming as many words as possible beginning with "P" and "S" within one minute. To date, 42 participants with MND enrolled on the study. Of these, 20 scored below 4 (indicating dysarthric speech) on the ALS Functional Rating Scale Revised (ALSFRS-R) \cite{cedarbaum1999alsfrs}, a clinician-administered scale assessing physical function and disability progression. Table \ref{tab:demo} summarises demographic characteristics of the participant cohort.

\begin{table}[ht]
\captionsetup{font=footnotesize}
\footnotesize 
\centering

\caption{Participants' demographic information, grouped by dysarthric (Dys) speech vs.\ typical (Typ).}
\label{tab:demo} 
\begin{tabular}{lccc} 
\toprule
\textbf{Characteristic} & \textbf{Dys (n=20)} & \textbf{Typ (n=22)} & \textbf{All (n=42)} \\
\midrule
Age  & 66.6 $\pm$ 13.1 & 63.0 $\pm$ 13.0 & 64.7 $\pm$ 13.0  \\
Female \% & 25.0 & 27.3  & 26.2\\
Education (years) & 14.0 $\pm$ 3.2 & 12.0 $\pm$ 2.4  & 13.0 $\pm$ 2.9
\\ 
ECAS Total & 112.7 $\pm$ 12.0 & 110.7 $\pm$ 14.8 & 110.8 $\pm$ 14.8\\
ALSFRS-R & 2.4 $\pm$ 0.8 & 3.6 $\pm$ 0.5  & 3.2 $\pm$ 1.0\\
VFI (P) & 4.7 $\pm$ 2.3 & 3.7 $\pm$ 1.7  & 4.2 $\pm$ 2.1\\
VFI (S) & 5.7 $\pm$ 3.6 & 4.0 $\pm$ 1.9  & 4.8 $\pm$ 3.0\\ 
\bottomrule
\end{tabular} 
\vspace{-0.4cm}
\end{table}

\subsection{Automatic Speech Recognition}
We evaluated several state-of-the-art automatic speech recognition (ASR) systems, including Wav2Vec2 \cite{baevski2020wav2vec2}, HuBERT \cite{hsu2021HuBERT}, NVIDIA NeMo \cite{nvidia2026nemo}, Whisper \cite{radford2023whisper}, and its extension, WhisperX \cite{bain2023whisperx}. WhisperX produced significantly more accurate word-level timestamps than the other systems, largely due to its phoneme-level forced alignment. Accordingly, we report results obtained using WhisperX in Section \ref{sec:results}. ASR performance was measured using the standard word error rate (WER) metric.

\vspace{-0.2cm}
\subsection{Voice Activity Detection}

To establish the most effective approach, we tested a range of VAD techniques on our dataset. Our investigation encompassed recent state-of-the-art voice activity detection toolkits, specifically Silero \cite{silero_vad}, FSMN (monophone-based) \cite{li2026fsmnmonophone}, and CNN-BiLSTM \cite{CNN_vad}. Silero VAD is a deep neural network-based model trained on a large-scale multilingual corpus exceeding 100 hours, offering strong generalisability and reliable speech/non-speech discrimination under adverse acoustic conditions. Segmentation performance was evaluated using metrics in VAD and diarisation tasks: speech false alarms (non-speech classified as speech), speech misses (speech classified as non-speech), pause false alarms (speech classified as pause), and pause misses (pause classified as speech). These four error types were summed and normalised by total audio duration to yield a weighted total segmentation error \cite{nist2004rt}.

\subsection{Clinical feature extractor}   
For linguistic analysis, we used SpaCy for text processing and named entity recognition (locations, people, organisations with GeoText), and NLTK's stemming to normalise word forms for consistent matching. These enabled correct word identification, repetition detection, and filler word flagging. Our feature extraction spanned three categories: (1) traditional acoustic, eGeMAPS (88 parameters), spectral descriptors, voice quality (jitter, shimmer), and articulation features; (2) self-supervised embeddings, Wav2Vec2 and HuBERT representations aggregated via mean and standard deviation; and (3) clinically interpretable, correct word count, average word duration, pause ratio, and disfluency counts. Table \ref{tab:feats} summarises these sets.

\begin{table}[ht] 
\vspace{-0.1cm}
\captionsetup{font=footnotesize}
\footnotesize
\centering
\caption{Summary of the extracted features.}
\label{tab:feats} 
\begin{tabular}{>{\raggedright\arraybackslash}p{1.9cm} p{2.2cm} p{2.7cm}}
\toprule
\textbf{Name} &  \textbf{Type (No.)} & \textbf{E.g.} \\
\midrule
eGeMaps & Acoustic (88) & Pitch, energy, intensity\\ 
eGeMaps (2) & Acoustic (175) & Stats of eGeMaps (Low-Level Descriptors)\\ 
Spectral/Temporal & Acoustic (43) & MFCCs, Vowel duration\\  
Formants & Acoustic (8) & F1, F2\\ 
Voice Quality & Acoustic (19) & Jitter, Shimmer\\

Articulation & Acoustic (9) & Formant trajectories, speech rates\\
\hline
Wav2Vec2 & Embedding (768) & Mean-last-layer\\  
HuBERT &  Embedding (1024) & Mean-layer (8,9)\\ 
\hline
Clinical &  Pause/Speech (31) & Stats of pauses, words, correct-words' length; pause/speech ratio; no of hesitations \\
\bottomrule
\end{tabular} 
\vspace{-0.3cm}
\end{table}   
 
\vspace{-0.2cm}
\subsection{Regression models} 
Our modelling approach explored both classical and deep learning methods. For traditional regression, we implemented Support Vector Regression (SVR), Extreme Gradient Boosting (XGB), Random Forest (RF), and LASSO. On the deep learning side, we attached regression heads to pre-trained transformers (Wav2Vec2, Whisper) and CNNs. Classical models consistently outperformed deep learning counterparts, which we attribute to our small dataset, where simpler models generalise better. Models were evaluated using 5-fold cross-validation with Normalised Root Mean Squared Error (NRMSE) and R². All models underwent hyperparameter tuning via grid search. For SVR, we optimised C [0.1, 1, 10] and kernel ['linear', 'rbf']. For LASSO, alpha was tuned across [0.001, 0.01, 0.1, 1]. For XGB, we explored n\_estimators [100, 300], max\_depth [3, 6], and learning\_rate [0.01, 0.1].
 \section{Results} 
\label{sec:results}
 \vspace{-0.1cm}
\subsection{ASR and adjustment performance}
For WhisperX (Large-V3), direct application performed suboptimally, so we explored two refinements: fine-tuning (weight updates on domain data) and prompting (task guidance, e.g., "words starting with a letter, excluding names"). Prompting provides contextual biasing, improving accuracy and rare word recognition for discrete fluency tasks. We first fine-tuned on our CognoSpeak corpus ($n>400$). For target tasks, fine-tuning reduced WER for P-words, while prompting worked better for S-words, likely due to P-word prevalence in training data and acoustic differences between plosives and fricatives. We excluded errors and compared performance across dysarthric versus normal speech and clinical versus control groups. Using 5-fold cross-validation (Table \ref{tab:wer}), WER was lower for P-words (21.5\%) than for S-words (23\%). Removing fillers improved performance by 14\% and 18.7\%, respectively. Dysarthric speakers had higher error rates, especially for S-words (typical: 15\%; dysarthric: 24\%).

\begin{table}[ht] 
\vspace{-0.2cm}
\captionsetup{font=footnotesize}
\footnotesize
\centering
\caption{WERs [\%] dysarthric (Dys) vs. Typical (Typ) speech. }
\label{tab:wer} 
\begin{tabular}{lcccc}
\toprule
\textbf{Task} & \textbf{Removed fillers} &   \textbf{Dys$\downarrow$} & \textbf{Typ$\downarrow$} &\textbf{All $\downarrow$}\\
\midrule
P & No & 28.8 & 17.1  & 21.5\\  
P & Yes & \textbf{22.2}  &  \textbf{8.1}   & \textbf{14.0}\\  
S & No & 37.1 & 12.8    & 23.0\\ 
S  & Yes &  24.0  & 15.0& 18.7\\
\bottomrule
\vspace{-0.4cm}
\end{tabular}  
\end{table}

Table \ref{tab:segerrors} presents segmentation errors for the three VAD systems following adjustments (longest and last segments). Silero demonstrated superior performance, yielding the lowest error rates among the systems evaluated: 6.9\% for P-words and 11.8\% for S-words.

\begin{table}[ht]
\captionsetup{font=footnotesize}
\footnotesize
\centering
\caption{Total segmentation (weighted speech/pause) errors.}
\label{tab:segerrors} 
\begin{tabular}{>{\raggedright\arraybackslash}p{2.25cm} p{1cm} p{1.5cm}  p{1.5cm}}
\toprule
\textbf{VAD} & \textbf{Adjust.} &   \textbf{P-words\%$\downarrow$} & \textbf{S-words \%$\downarrow$}\\
\midrule
Silero \cite{silero_vad}& Longest  &  \textbf{6.9}  &  \textbf{11.8} \\ 
CNN-BiLSTM \cite {CNN_vad}& Last &  11.0  &  13.5\\ 
FSMN \cite{li2026fsmnmonophone}& Longest  &  15.6  &  20.7 \\  

\bottomrule 
\end{tabular} 
\end{table}

 \vspace{-0.5cm}
\subsection{Regression models evaluation} 
Table \ref{tab:vfi-oracle} shows the VFI estimation results of the two best regression models for the three experiments: the oracle condition used clinical features derived from manual alignments (Clin-Man), establishing an upper performance bound, clinical features extracted from the automatic alignments (Clin-Man), and the baseline (Base) condition comprised models trained on non-clinical features, providing a point of comparison to assess the value added by clinical features. For the P-words, LASSO emerged as the best-performing model in the oracle condition, achieving an NRMSE of 0.064 and an R² of 0.797. Similarly, for the S-words, LASSO achieved an NRMSE of 0.066 and a strong R² of 0.865. These results demonstrate that, under oracle conditions, both word sets enable highly accurate VFI estimation, closely approximating the true values. In contrast, the best-performing baseline model, SVR trained on HuBERT layer 8, produced substantially worse results. For P-words, this model achieved an NRMSE of 0.102 and an R² of 0.526; for S-words, performance degraded further to an NRMSE of 0.164 and an R² of 0.353. These results are significantly inferior to those obtained under oracle conditions. It is worth noting that models trained on embeddings surpassed those relying on other non-clinical feature types. However, using our final fully automatic pipeline, LASSO and SVR models performed best for both P and S-words. For P-words, LASSO achieved strong results (R² 0.856, NRMSE 0.053), comparable to or even better than oracle conditions (most of the points are close to the $y=x$ dashed line). For S-words, performance degraded (R² 0.832, NRMSE 0.083). We found no statistically significant difference between LASSO models trained on Clin-Man and Clin-Auto features for both target words (paired t-test on results from 10× 5-fold cross-validation with different random seeds).

\begin{table}[ht]
\captionsetup{font=footnotesize}
\footnotesize
\centering
\caption{VFI estimation results for the top two models trained on manual clinical features vs baseline features. (Average/STD).}
\label{tab:vfi-oracle}  
\begin{tabular}{>{\raggedright\arraybackslash}p{0.7cm} p{2.3cm} p{1.5cm}  p{1.5cm}}
\toprule
\textbf{Model} &  \textbf{Feat.} & \textbf{R²$\uparrow$}& \textbf{NRMSE$\downarrow$} \\ 
\midrule
LASSO & Clin-Man (P) & 0.797/0.131& 0.064/0.024\\  
SVR & Clin-Man (P) & 0.793/0.124 & 0.067/0.033\\ 
\hline
LASSO & Clin-Auto (P) &  \textbf{0.856/0.111}  & \textbf{0.053/0.015}  \\  
SVR & Clin-Auto (P) &   0.808/0.135  & 0.065/0.028  \\ 

\hline
SVR & Base-HuBERT-8 (P) &0.526/0.190& 0.102/0.008\\  
SVR & Base-HuBERT-9 (P) & 0.497/0.210 & 0.105/0.009\\ 

\midrule
LASSO & Clin-Man (S) & 0.865/0.106& \textbf{0.066/0.034}\\  

SVR & Clin-Man (S) &  \textbf{0.873/0.071}  & 0.070/0.037 \\ 
\hline
LASSO & Clin-Auto (S) &  0.832/0.041 & 0.083/0.042  \\  
SVR & Clin-Auto (S) & 0.741/0.135  & 0.102/0.060 \\ 
\hline
SVR & Base-HuBERT-8 (S) &0.353/0.342& 0.164/0.014\\  
RF & Base-Wav2Vec2 (S) &0.060/0.926 & 0.168/0.014\\ 
\bottomrule
\end{tabular}  
\vspace{-0.2cm}
\end{table}





 
 
\begin{figure}[t]  
\captionsetup{font=footnotesize}
\footnotesize 
  \includegraphics[width=1.1\linewidth]{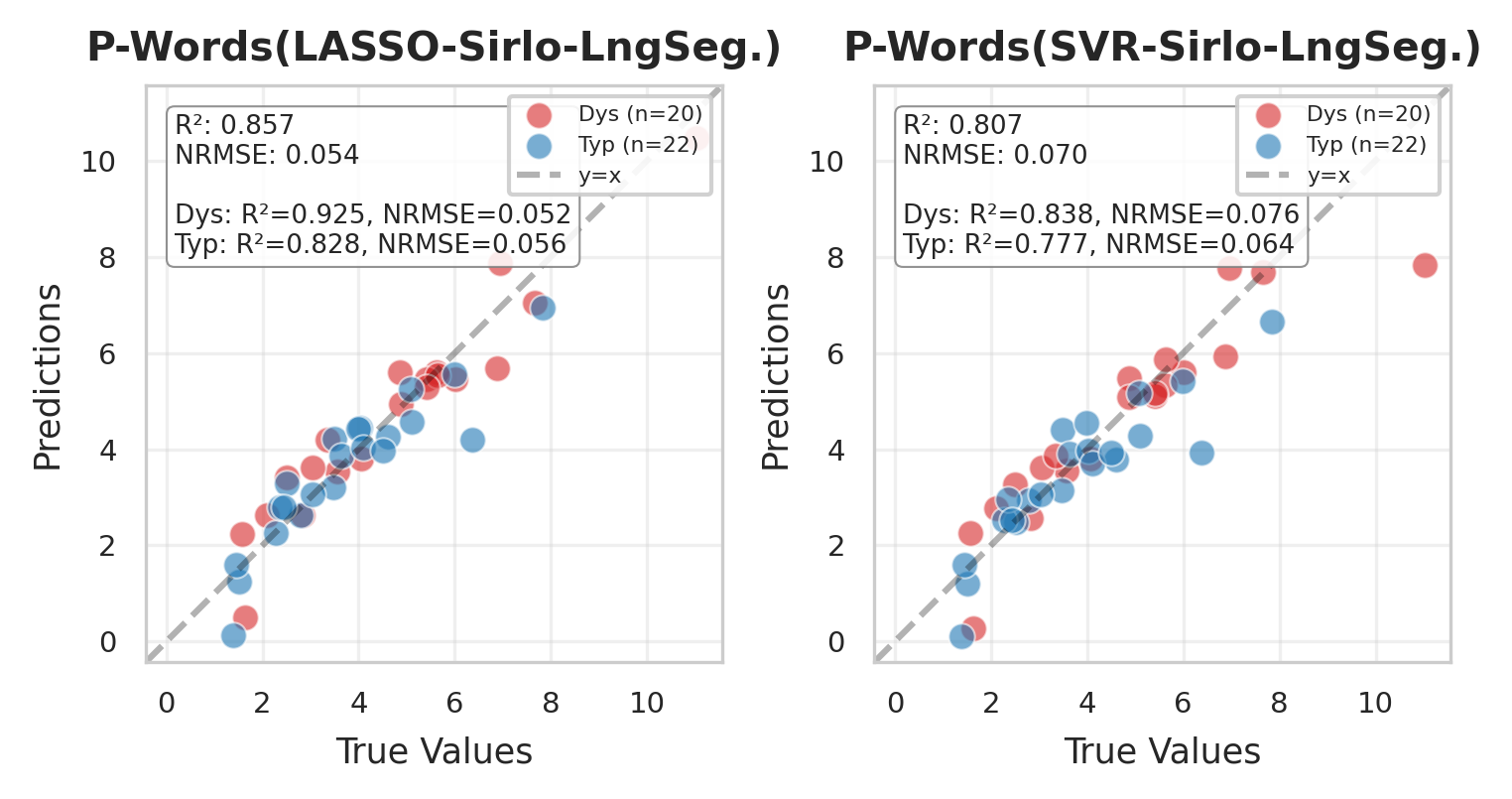}
  
    \includegraphics[width=1.1\linewidth]{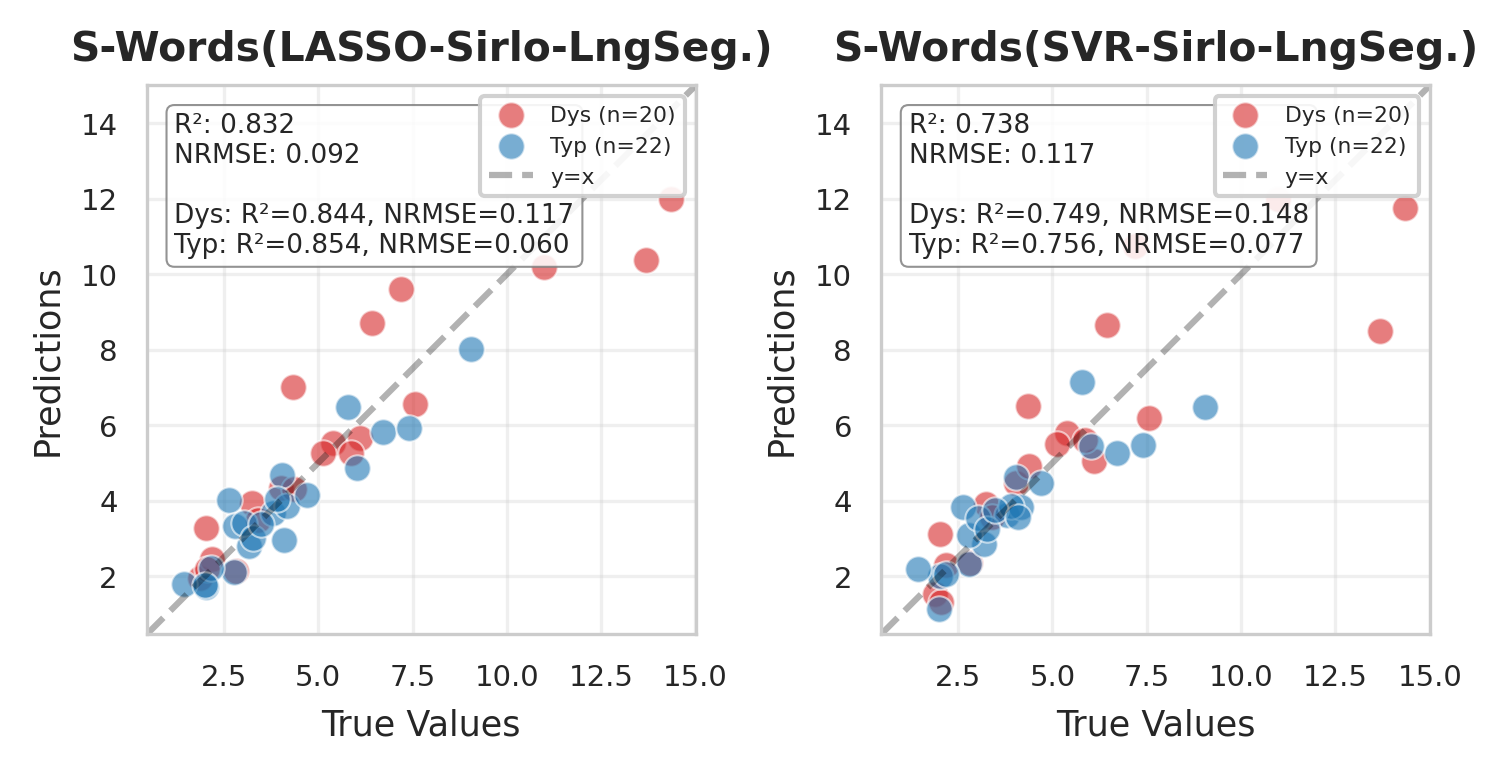}
  \caption{Scatter plots for the best 2 models (top: P-words, bottom: S-words)}
  \label{fig:scatter} 
  \vspace{-0.6cm}
\end{figure}

Scatter plots of estimations from the two top models were plotted using data aggregated across all five test folds. estimation errors were similar for dysarthric and typical speakers (Figure 3), indicating a strong generalisation. A substantial group disparity emerged: NRMSE was 0.060 for typical speakers versus 0.117 for dysarthric speakers, suggesting that higher ASR word error rates and segmentation inaccuracies disproportionately affected dysarthric speech, reducing estimation accuracy for S-words. Note that some VFIs, particularly for dysarthric speakers, exceeded 12, resulting in more errors seen in the top-right region of the figures.

\section{Conclusions}
\label{sec:conclusions} 

This study establishes the feasibility of an automated pipeline that integrates state-of-the-art ASR and VAD with interpretable clinical features to generate accurate regression models for VFI estimation. Although the standard ECAS fluency assessment employs S-words, we demonstrate that P-words yield comparable estimation performance, with estimation achievable using only the generation phase. Future work should incorporate supplementary tasks (e.g., reading paragraphs, picture descriptions), particularly for dysarthric populations who exhibit elevated WERs.  
\clearpage
\section{Generative AI Use Disclosure}
During the preparation of this manuscript, generative AI tools were employed solely for language editing, clarity refinement, and paragraph summarisation. All aspects of the study, including design, data collection, experiments, analysis, and conclusions, were conducted entirely by the authors. AI use was confined to improving readability, grammar, and formatting, with no influence on experimental outcomes or their interpretation.

\bibliographystyle{IEEEtran}
\bibliography{mybib} 
\end{document}